\documentclass{article}

\usepackage{makecell}
\usepackage{microtype}
\usepackage{graphicx}
\usepackage{subfigure}
\usepackage{booktabs} % for professional tables
\usepackage[dvipsnames]{xcolor}         % code listings colors

\usepackage[colorlinks=true,citecolor=Blue,linkcolor=black]{hyperref}
\usepackage[accepted]{mlsys2020}
\usepackage{url}            % simple URL typesetting
\usepackage{amsfonts}       % blackboard math symbols
\usepackage{nicefrac}       % compact symbols for 1/2, etc.
\usepackage{microtype}      % microtypography
\usepackage{listings}       % code listings

\usepackage{xcolor}

\makeatletter
\let\old@lstKV@SwitchCases\lstKV@SwitchCases
\def\lstKV@SwitchCases#1#2#3{}
\makeatother
\usepackage{lstlinebgrd}
\makeatletter
\let\lstKV@SwitchCases\old@lstKV@SwitchCases

\lst@Key{numbers}{none}{%
    \def\lst@PlaceNumber{\lst@linebgrd}%
    \lstKV@SwitchCases{#1}%
    {none:\\%
     left:\def\lst@PlaceNumber{\llap{\normalfont
                \lst@numberstyle{\thelstnumber}\kern\lst@numbersep}\lst@linebgrd}\\%
     right:\def\lst@PlaceNumber{\rlap{\normalfont
                \kern\linewidth \kern\lst@numbersep
                \lst@numberstyle{\thelstnumber}}\lst@linebgrd}%
    }{\PackageError{Listings}{Numbers #1 unknown}\@ehc}}
\makeatother

\definecolor{codegreen}{rgb}{0,0.6,0}
\definecolor{codegray}{rgb}{0.5,0.5,0.5}
\definecolor{codepurple}{rgb}{0.58,0,0.82}
\definecolor{backcolor}{rgb}{0.95,0.95,0.95}
\definecolor{keywordcolor}{rgb}{0, 0.1, 0.8}
\definecolor{ivycolor}{rgb}{0, 0.5, 0}
\definecolor{pinkcolor}{rgb}{0.75, 0.22, 0.17}
\definecolor{backendcolor}{RGB}{255,150,50}
\definecolor{ivyeagercolor}{RGB}{100,150,100}
\definecolor{backendbackcolor}{RGB}{255,200,100}
\definecolor{ivyeagerbackcolor}{RGB}{150,200,150}
\definecolor{ivycompilablebackcolor}{RGB}{50,150,50}

\lstdefinestyle{mystyle}{
    language=Python,
    backgroundcolor=\color{backcolor},   
    commentstyle=\color{codegray},
    keywordstyle={\color{keywordcolor}},
    keywordstyle = [2]{\color{keywordcolor}},
    keywordstyle = [3]{\color{pinkcolor}},
    keywordstyle = [4]{\color{pinkcolor}},
    keywordstyle = [5]{\color{ivycolor}},
    keywordstyle = [6]{\color{ivycolor}},
    keywordstyle = [7]{\color{ivycolor}},
    keywordstyle = [8]{\color{ivycolor}},
    keywordstyle = [9]{\color{ivycolor}},
    keywordstyle = [10]{\color{ivycolor}},
    keywordstyle = [11]{\color{ivycolor}},  
    otherkeywords = {None},
    morekeywords = [2]{None},
    morekeywords = [3]{__init__},
    morekeywords = [4]{self},
    morekeywords = [5]{ivy},
    morekeywords = [6]{ivy_vision},
    morekeywords = [7]{ivy_robot},
    morekeywords = [8]{ivy_mech},
    morekeywords = [9]{ivy_gym},
    morekeywords = [10]{f},
    morekeywords = [11]{_f},
    numberstyle=\tiny\color{codegray},
    stringstyle=\color{codepurple},
    basicstyle=\ttfamily\footnotesize,
    breakatwhitespace=false,         
    breaklines=true,                 
    captionpos=b,
    columns=fullflexible,
    keepspaces=true,                 
    numbers=left,                    
    numbersep=5pt,                  
    showspaces=false,                
    showstringspaces=false,
    showtabs=false,                  
    tabsize=2
}

\begin{document}

\twocolumn[
\mlsystitle{Ivy: Templated Deep Learning for Inter-Framework Portability}

% unless you've provided the [accepted] option to the mlsys2020


\mlsyssetsymbol{equal}{*}

\begin{mlsysauthorlist}
\mlsysauthor{Daniel Lenton}{icl}
\mlsysauthor{Fabio Pardo}{icl}
\mlsysauthor{Fabian Falck}{ox}
\mlsysauthor{Stephen James}{icl}
\mlsysauthor{Ronald Clark}{icl}
\end{mlsysauthorlist}

\mlsysaffiliation{icl}{Department of Computing, Imperial College London, UK}
\mlsysaffiliation{ox}{University of Oxford, UK}

\mlsyscorrespondingauthor{Daniel Lenton}{djl11@ic.ac.uk}

\mlsyskeywords{Machine Learning, MLSys}

\vskip 0.3in

\begin{abstract}

We introduce Ivy, a templated Deep Learning (DL) framework which abstracts existing DL frameworks. Ivy unifies the core functions of these frameworks to exhibit consistent call signatures, syntax and input-output behaviour. New high-level framework-agnostic functions and classes, which are usable alongside framework-specific code, can then be implemented as compositions of the unified low-level Ivy functions. Ivy currently supports TensorFlow, PyTorch, MXNet, Jax and NumPy. We also release four pure-Ivy libraries for mechanics, 3D vision, robotics, and differentiable environments. Through our evaluations, we show that Ivy can significantly reduce lines of code with a runtime overhead of less than 1\% in most cases. We welcome developers to join the Ivy community by writing their own functions, layers and libraries in Ivy, maximizing their audience and helping to accelerate DL research through inter-framework codebases. More information can be found at \url{https://ivy-dl.org}.
\end{abstract}
]



\printAffiliationsAndNotice{}  % leave blank if no need to mention equal contribution
% \printAffiliationsAndNotice{\mlsysEqualContribution} % otherwise use the standard text.

\section{Introduction}
\label{sec:intro}

%\subsection{Hierarchy of Abstractions}

There is generally a trade-off in software projects between run-time efficiency and ease of development. At a high level, this trade-off is intuitive; programming solutions with more abstractions remove complexity, but also necessarily remove control, and the ability to perform task-specific optimizations. Effective frameworks must find a middle ground between these two competing factors, where the right abstractions are needed to make development as quick and easy as possible, whilst also enabling customized implementations for maximum runtime efficiency and control.

In the context of Deep Learning (DL) frameworks, Python has emerged as the front-runner language for research and development. Most DL frameworks depend on efficient pre-compiled C++ code in the backend, which is a clear example of finding an effective balance between these competing factors. The Python interface makes prototyping code quick and easy, and the pre-compiled C++ operations and CUDA kernels in the backend make model inference fast. While users of most DL frameworks are still given the option for C++ and CUDA development of custom operations, the most common use case is for developers to implement their projects as compositions of operations in pure Python. The abstractions available for this development style also continue to become more powerful. For example, most frameworks now enable chains of Python functions to be flagged for Just-In-Time (JIT) compilation, using tools such as the Accelerated Linear Algebra compiler (XLA) \cite{leary2017xla}.

\begin{figure}[h!]
\centering
\includegraphics[width=0.47\textwidth]{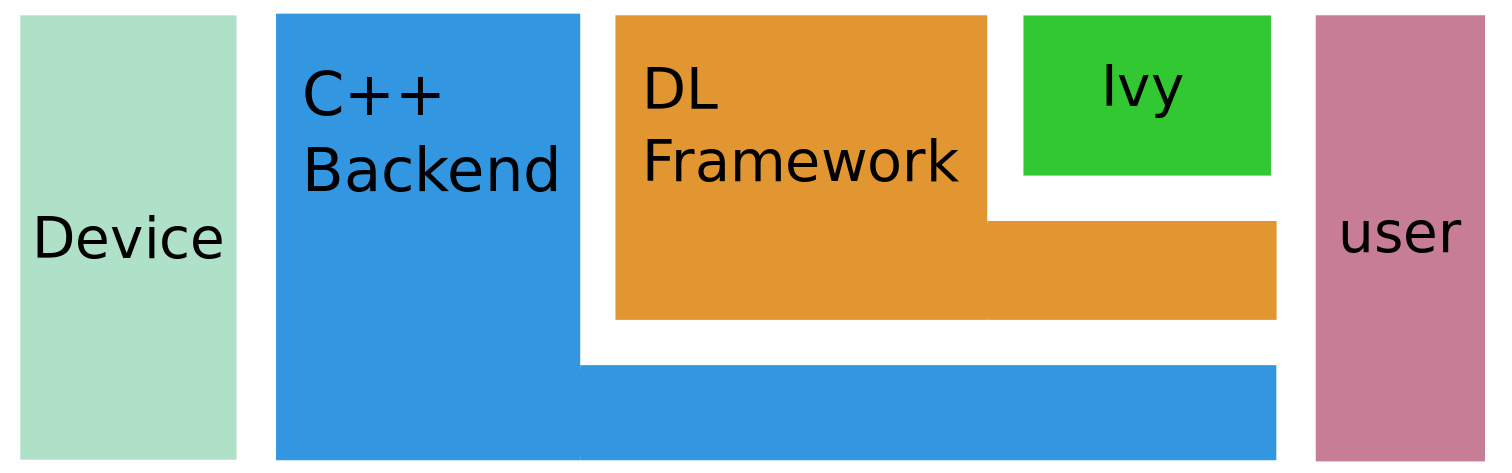}
  \caption{A simplified schema illustrating how Ivy sits above existing DL frameworks in the abstraction hierarchy, with the C++ backend sitting below the DL framework. All levels of abstraction remain accessible to the developer, allowing maximal control.}
  \label{fig:abstraction_hierarchy}
\end{figure}

We frame Ivy in the same hierarchy of abstractions (see Figure \ref{fig:abstraction_hierarchy}). Ivy abstracts existing DL frameworks such that their functional Application Programming Interfaces (APIs) all exhibit consistent call signatures, syntax and input-output behaviour. In doing so, Ivy effectively moves existing DL frameworks one layer down the abstraction stack to the Ivy ``backend''. As with the abstracted C++ backend in DL frameworks, we find the benefits of the Ivy abstraction generally outweigh the costs. New functions written in Ivy are instantly portable to TensorFlow, PyTorch, MXNet, Jax, and NumPy, enabling an inter-framework ``drag-and-drop'' approach not currently possible among modern DL frameworks. If a new Python DL framework was introduced in future, adding this framework to the Ivy backend would then make all existing Ivy code instantly compatible with the new framework. Ivy offers the potential for creating ``lifelong'' framework-agnostic DL libraries, which are jointly usable by present and future DL developers in present and future frameworks.

\subsection{Towards General Differentiable Programming}
\label{sec:gdp}

Although DL initially focused on end-to-end training of deep neural networks (DNNs), DL models increasingly use a hybrid of neural networks and parameter-free, ``hand-designed'' components that encode priors and domain-specific knowledge from the relevant field \cite{software2}. Robotic control, path planning and Structure from Motion (SfM) are just a few examples. Most of these fields have very well-established mathematical foundations which pre-date DL. The more successful intersections with DL usually find an effective middle ground where known parameter-free functions can still be exploited in the end-to-end computation graph. The only requirement is that these parameter-free computation blocks can still pass gradients for the end-to-end learning.

We show an example of using a parameter-free function from the Ivy vision library in a TensorFlow neural network model below. The model receives a color image \lstinline[style=mystyle]{rgb} and corresponding 3D co-ordinates \lstinline[style=mystyle]{coords}, encodes features from \lstinline[style=mystyle]{rgb} via a 2D convolution, and then uses \lstinline[style=mystyle]{coords} to construct a 3D voxel gird of these features, which is then further processed by 3D convolutions for reasoning about the 3D scene. This examples demonstrates the supplementary nature of Ivy functions, which can be used alongside native frameworks, TensorFlow in this case. The real power of Ivy is that the function on ln 15 - 16 can be used as is in \textit{any supported framework} (i.e. PyTorch, Jax, etc.).

% \begin{minipage}{\linewidth}
\begin{lstlisting}[
style=mystyle,
label=ivy_lib_user_code,
breakindent=40pt]
import tensorflow as tf
from tensorflow.keras.layers import Layer, Conv2D, Conv3D
import ivy_vision

class TfModel(Layer):
  def __init__(self):
    super().__init__()
    self._conv2d = Conv2D(16, 3)
    self._conv3d = Conv3D(1, 3)
    
  def call(self, coords, rgb):
    feat = self._conv2d(rgb)
    fs = feat.shape
    feat = tf.reshape(feat, (fs[0], fs[1]*fs[2], fs[3]))
    vox = ivy_vision.coords_to_voxel_grid(
        coords, [128] * 3, features=feat)
    return self._conv3d(vox[0])
\end{lstlisting}
% \end{minipage}

These types of differentiable domain-specific functions are becoming ever more ubiquitous in deep learning research. One of the most prominent fields to combine prior knowledge with end-to-end learning is computer vision. Indeed, the convolutional architecture itself \cite{lecun1989backpropagation} is an example of inductive bias in the computation graph, driven by a heuristic of local spatial significance in images. More recent works in computer vision have incorporated well-known multi-view geometry relations into the graph, which can greatly help in establishing correspondence between images. FlowNet \cite{dosovitskiy2015flownet} shows that adding explicit correlations over image patches greatly improves correspondence estimation over vanilla CNNs. 
Many works which combine DL with SfM for geometric reconstructions also utilize core image projection and warping functions in the graph \cite{tang2018ba, bloesch2018codeslam}, again requiring gradient propagation.

Gradient based optimization also pre-dates DL in many applied fields, such as motion planning. Works such as CHOMP \cite{ratliff2009chomp} and TrajOpt \cite{schulman2014motion} demonstrate that motion planning can be done through gradient-based optimization. More recently, path planning has seen interesting intersections with DL. For example, Value Iteration Networks (VIN) \cite{tamar2016value} utilize the value-iteration structure for ``learning to plan''.

Outside of robotics and computer vision, other fields are increasingly exploiting parameter-free computation in end-to-end graphs. \citep{raissi2020hidden} propose a physics-informed deep-learning framework capable of encoding the Navier-Stokes equations into neural networks with applications in Fluid Mechanics, \citep{graves2014neural, sukhbaatar2015end} learn to solve memory intensive tasks from data by integrating differentiable read and write operations into a neural network with an external memory bank, and \citep{qiao2020scalable} propose a differentiable physics framework which uses meshes and exploits the sparsity of contacts for scalable differentiable collision handling.

These are just some examples of the growing need for libraries which provide domain specific functions with support for gradient propagation, to enable their incorporation into wider end-to-end pipelines. We provide an initial set of Ivy libraries for mechanics, 3D vision, robotics, and differentiable environments. We expect these initial libraries to be widely useful to researchers in applied DL for computer vision and robotics. We explore these libraries further in Section \ref{sec:ivy_libraries}, and provide an end-to-end example in Section \ref{sec:end-to-end}.

% \begin{figure}[h!]
% \centering
% \includegraphics[width=0.48\textwidth]{figures/computation_graph.png}
%   \caption{Example computation graph for robotic visuomotor control, involving both learnt parameter-based and non-learnt parameter-free computation. Ivy is particularly well suited for abstracting domain-specific parameter-free functions.}
%   \label{fig:computation_graph}
% \end{figure}

\subsection{A Templated Framework}

In order to abstract DL frameworks, Ivy takes inspiration from the concepts of template metaprogramming \cite{abrahams2004c++} and template methods \cite{gamma1995design}. Template metaprogramming refers to compile-time polymorphism, enabling source code to compile against different data types, while template methods are a behavioral design pattern for object oriented programming, reducing lines of code by delegating low-level implementations of general abstract functions to more specific child classes. The template concept remains similar in both cases, allowing the creation of individual functions which can take on a variety of forms at runtime. Ivy takes inspiration from this general concept, and uses templates at the level of DL frameworks.

%When looking at how DL frameworks have evolved over the past few years, a general trend is evident: Increasingly powerful abstractions, which allow developers to implement code in higher level languages aimed at rapid prototyping, have significantly reduced the time investment for highly-efficient DL solutions. This has undoubtedly contributed to the increasingly rapid dispersal of new ideas in the field. With Ivy, we continue the trend of increased abstraction, but from a different angle.

Ivy enables functions, layers and libraries to be implemented once, with simultaneous support for all prominent modern Python DL frameworks. Unlike Keras \cite{chollet2015keras}, we do not attempt to abstract high level classes. Aside from this being more difficult to maintain, this level of abstraction removes control from users. Instead, we abstract only the core tensor functions, which are often semantically similar, but syntactically unique.

This design enables functions in all Ivy libraries to be ``dragged and dropped'' into any project using a supported framework. We will continue to expand Ivy's applied libraries, and we encourage users to join the Ivy community by implementing their own functions, layers and libraries in Ivy to maximize their audience, and help accelerate DL research through the creation of inter-framework codebases.
\section{Related Work}

\subsection{Deep Learning Frameworks}

Deep learning progress has evolved rapidly over the past decade, and this has spurred companies and developers to strive for framework supremacy. Large matrix and tensor operations underpin all efficient DL implementations, and so there is largely more that relates these frameworks than separates them. Many frameworks were designed explicitly for matrix and tensor operations long before the advent of modern DL. In the Python language \cite{10.5555/1593511}, one of the most widely used packages is NumPy \cite{oliphant2006guide, harris2020array}, which established itself as a standard in scientific computing. NumPy is a general matrix library, but with many function implementations highly optimized in C \cite{kernighan2006c}. It does not natively support automatic differentiation and back-propagation. Since the beginning of the new DL era, a number of libraries with automatic differentiation have been utilized. An early and widely used library was Caffe \cite{jia2014caffe}, written in C++ \cite{stroustrup2000c++}, enabling static graph compilation and efficient inference. The Microsoft Cognitive Toolkit (CNTK) \cite{seide2016cntk} was also written in C++, and supported directed graphs. Both of these are now deprecated. More recently, Python has become the front-runner language for DL interfaces. TensorFlow \cite{tensorflow2015-whitepaper}, Theano \cite{2016arXiv160502688short}, Chainer \cite{tokui2019chainer}, MXNet \cite{chen2015mxnet}, PyTorch \cite{paszke2019pytorch} and JAX \cite{jax2018github} are all examples of DL frameworks primarily for Python development.

Despite the variety in frameworks, the set of fundamental tensor operations remains finite and well defined, and this is reflected in the semantic consistency between the core tensor APIs of all modern python DL libraries, which closely resemble that of NumPy introduced in 2006. Ivy abstracts these core tensor APIs, with scope to also abstract future frameworks adhering to the same pattern, offering the potential for inter-framework code reusability long into the future.

\subsection{Deep Learning Libraries}

Many field-specific libraries exist, for example DLTK \cite{pawlowski2017state} provides a TensorFlow toolkit for medical image analysis, PyTorch3D \cite{ravi2020pytorch3d} implements a library for DL with 3D data, PyTorch Geometric \cite{fey2019fast} provides methods for deep learning on graphs and other irregular structures, and ZhuSuan \cite{zhusuan2017} is a TensorFlow library designed for Bayesian DL. Officially supported framework extensions are also becoming common, such as GluonCV and GluonNLP \cite{gluoncvnlp2020} for MXNet, TensorFlow Graphics \cite{TensorflowGraphicsIO2019}, Probability \cite{dillon2017tensorflow}, and Quantum \cite{broughton2020tensorflow} for TensorFlow, and torchvision and torchtext for PyTorch \cite{paszke2019pytorch}. However, the frameworks these libraries are built for can quickly become obsoleted, also making the libraries obsolete. Furthermore, none of these libraries address the code shareability barrier for researchers working in different frameworks. As of yet, there is no solution for building large framework-agnostic libraries for users of both present and future DL Python frameworks. Ivy offers this solution. 

\subsection{Deep Learning Abstractions}

Some previous works do provide framework-level abstractions for DL. Keras \cite{chollet2015keras} supported TensorFlow \cite{tensorflow2015-whitepaper}, CNTK \cite{seide2016cntk}, and Theano \cite{2016arXiv160502688short} before it's focus shifted to support TensorFlow only. Keras provided abstractions at the level of classes and models, enabling users to prototype quickly with higher level objects and standard training pipelines. However, for researchers adopting non-standard pipelines, this level of abstraction can remove too much control. In contrast, Ivy simplifies and reduces the abstraction to the core tensor APIs, enabling new libraries and layers to be built on top of Ivy's functional API in a highly scalable, maintainable and customized manner.

TensorLy \cite{kossaifi2016tensorly} is closer in scope to Ivy, also offering a framework agnostic functional API. However, many functions such as gather, scatter and neural network functions such as convolutions are not provided by TensorLy. The library instead focuses on general tensor operations such as decomposition, regression and sparse tensor handling. The TensorLy backend focuses on the functions necessary to support these operations. Ivy offers a much more comprehensive API, applicable to a wide variety of fields, which is showcased in the applied libraries for mechanics, vision, robotics, and differentiable environments.
\section{Ivy Core}
\label{sec:IvyCore}

We now provide an overview of the core Ivy API and explain how backend frameworks are selected by the user and managed internally. All Ivy functions are unit tested against each backend framework, and support arbitrary batch dimensions of the inputs. The existing core functions are sufficient for implementing a variety of examples through the four Ivy applied libraries, but the core Ivy API can easily be extended to include additional functions.

\subsection{Framework-Specific Namespaces}

Almost all of the functions in the core Ivy API exist in the native frameworks in some form. Ivy wraps these native functions to provide consistent syntax and call signatures, and in some cases also extend functionality to achieve this goal. This is necessary in cases where the native functions are lacking, for example \lstinline[style=mystyle]{ivy.torch.gather_nd} is implemented by wrapping the less general \lstinline[style=mystyle]{torch.gather}. In general, the input-output behaviour for each Ivy function is selected to be the most general variant among the backends, whilst following the most common syntax.

\begin{figure}[h!]
\centering
\includegraphics[width=0.5\textwidth]{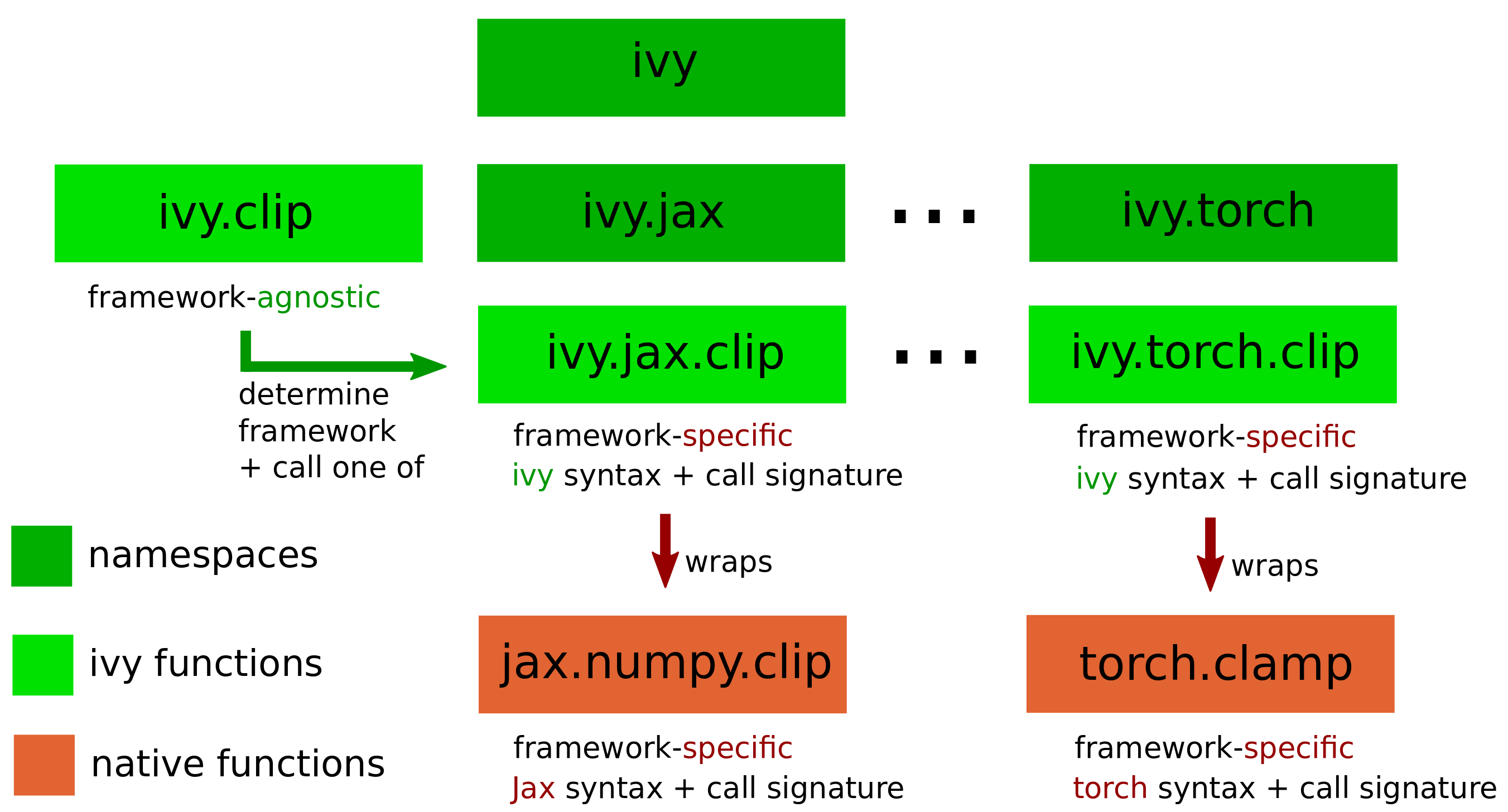}
  \caption{Overview of the core Ivy API.}
  \label{fig:namespaces}
\end{figure}

The framework-specific functions with the updated Ivy syntax and call signatures are all accessible via framework-specific namespaces such as \lstinline[style=mystyle]{ivy.tensorflow} and \lstinline[style=mystyle]{ivy.torch}, see Figure \ref{fig:namespaces}. Each of these namespaces behave like the functional API of the original framework, but with the necessary changes to bring inter-framework unification.

Due to the semantic similarity between all DL frameworks, these changes are very minor for most functions, with many changes being purely syntactic, which enables direct bindings to the native functions. Other functions require simple re-arrangement of the arguments, and sometimes extra processing of optional arguments to unify default behaviour. We show how Ivy wraps PyTorch functions with varying extents of modification below. A full runtime analysis of the Ivy overhead for each core function averaged across the backend frameworks is given in Section \ref{sec:RuntimeAnalysis}, and framework-specific overheads are provided in Appendix \ref{app:runtime_analysis}.

% \begin{minipage}{\linewidth}
\begin{lstlisting}[
style=mystyle,
label=container_config]
# direct binding
clip = torch.clamp

# minimal change
tile = lambda x, reps: x.repeat(reps)

# moderate change
def cast(x, dtype_str):
  dtype_val = torch.__dict__[dtype_str]
  return x.type(dtype_val)
    
# larger change
def transpose(x, axes=None):
  if axes is None:
    axes = range(len(x.shape)-1, -1, -1)
  return x.permute(axes)
\end{lstlisting}
% \end{minipage}

\subsection{Framework-Agnostic Namespace}
\label{sec:framework_agnostic}

While the framework-specific namespaces do provide inter-framework unification of syntax and call signatures, each of these namespaces are still \textit{specific} to a single framework. In order to create new framework \textit{agnostic} code, a framework-agnostic Ivy namespace is used, with functions accessible directly as \lstinline[style=mystyle]{ivy.func_name}. Every function in the framework-specific namespaces is also implemented in the framework-agnostic namespace, in exactly two lines of code like so: 

% \begin{minipage}{\linewidth}
\begin{lstlisting}[
style=mystyle,
% caption=Global framework setting, 
label=global_framework_setting]
def clip(x, x_min, x_max, f=None):
  return ivy.get_framework(x, f=f).clip(x, x_min, x_max)
    
def some_fn(*args, f=None):
  return ivy.get_framework(*args, f=f).some_fn(*args)
\end{lstlisting}
% \end{minipage}

Each function makes use of the method \lstinline[style=mystyle]{get_framework(*args, f=f)}, which returns the desired framework-specific namespace such as \lstinline[style=mystyle]{ivy.tensorflow} or  \lstinline[style=mystyle]{ivy.torch}. The bound framework-specific method is then directly called.

Some Ivy functions are only available in the framework-agnostic namespace, such as \lstinline[style=mystyle]{ivy.lstm_update}. These are implemented as compositions of other lower level functions.

With these framework-agnostic functions, new framework-agnostic code can then easily be implemented by composing them in new ways. A remaining question is, how does the user specify which backend framework to use at runtime?

\subsection{Local Framework Specification}

Local framework specification allows the backend to be specified on a per-function basis. In this mode, the framework-agnostic functions as outlined above are called directly, and the method \lstinline[style=mystyle]{get_framework(*args, f=f)} is responsible for selecting the backend. \lstinline[style=mystyle]{get_framework(*args, f=f)} determines the desired backend using one of two possible mechanisms.

\paragraph{Framework argument} The framework can be specified for any core function call using the \lstinline[style=mystyle]{f} argument, which can either be specified as a string such as \lstinline[style=mystyle]{'tensorflow'} or as the backend namespace itself such as \lstinline[style=mystyle]{ivy.tensorflow}. When \lstinline[style=mystyle]{f} is a string, a dictionary lookup is used, and when \lstinline[style=mystyle]{f} is a namespace, \lstinline[style=mystyle]{f} is returned unmodified. If \lstinline[style=mystyle]{f} is specified, this always takes priority for backend selection.

% \begin{minipage}{\linewidth}
\begin{lstlisting}[
style=mystyle,
% caption=Framework argument, 
label=framework_argument]
x = tf.constant([-1, 2.])
y = ivy.clip(x, 0, 1, f='tensorflow')
y = ivy.clip(x, 0.2, 0.8, f=ivy.tensorflow)
x = torch.tensor([-1, 2.])
y = ivy.clip(x, 0, 1, f='torch')
y = ivy.clip(x, 0.2, 0.8, f=ivy.torch)
\end{lstlisting}
% \end{minipage}

\paragraph{Type checking} Rather than declaring the \lstinline[style=mystyle]{f} argument for every function call, the correct framework can automatically be inferred by type checking of the inputs. To avoid importing all of the supported native frameworks for type checking, the types of the input arguments are instead converted to strings for specific keywords search. Importantly, this prevents the need to have all supported native frameworks installed locally just for type-checking. Type-checking is more user-friendly than specifying \lstinline[style=mystyle]{f} for every function call, but the continual type-checking adds a small runtime overhead when running in eager mode. If \lstinline[style=mystyle]{f} is specified, this takes priority, and type checking of the inputs is not used.

% \begin{minipage}{\linewidth}
\begin{lstlisting}[
style=mystyle,
% caption=Type Checking, 
label=type_checking]
x = tf.constant([-1, 2.])
y = ivy.clip(x, 0, 1)
x = torch.tensor([-1, 2.])
y = ivy.clip(x, 0, 1)
\end{lstlisting}
% \end{minipage}

\subsection{Global Framework Setting}

A framework can also be set globally for all future function calls using the method \lstinline[style=mystyle]{ivy.set_framework()}. Again, the framework can be specified either as a string or as the backend namespace. The backend can then be unset using \lstinline[style=mystyle]{ivy.unset_framework()}. With most projects using a single framework at runtime, this is the recommended framework selection mode.

% \begin{minipage}{\linewidth}
\begin{lstlisting}[
style=mystyle,
% caption=Global Framework Setting, 
label=global_framework_setting]
x = tf.constant([-1, 2.])
ivy.set_framework('tensorflow')
y = ivy.clip(x, 0, 1)
y = ivy.clip(x, 0.2, 0.8)
ivy.unset_framework()
x = torch.tensor([-1, 2.])
ivy.set_framework(ivy.torch)
y = ivy.clip(x, 0, 1)
y = ivy.clip(x, 0.2, 0.8)
ivy.unset_framework()
\end{lstlisting}
% \end{minipage}

The framework can also be set globally for a block of code by using the \lstinline[style=mystyle]{with} command like so:

% \begin{minipage}{\linewidth}
\begin{lstlisting}[
style=mystyle,
% caption=Global Framework Block Setting, 
label=global_framework_block_setting]
x = tf.constant([-1, 2.])
with ivy.tensorflow.use:
  y = ivy.clip(x, 0, 1)
  y = ivy.clip(x, 0.2, 0.8)
x = torch.tensor([-1, 2.])
with ivy.torch.use:
  y = ivy.clip(x, 0, 1)
  y = ivy.clip(x, 0.2, 0.8)
\end{lstlisting}
% \end{minipage}

when setting and unsetting globally, the \lstinline[style=mystyle]{__dict__} attribute of the framework-specific namespace, such as \lstinline[style=mystyle]{ivy.tensorflow}, is iterated, and each key and value pair is used to directly replace the same key and value pair of the framework-agnostic Ivy namespace \lstinline[style=mystyle]{ivy}. This entirely bypasses the framework-agnostic functions as outlined in section \ref{sec:framework_agnostic}, and instead binds all methods of the framework-specific namespace directly to the framework-agnostic namespace. Methods such as \lstinline[style=mystyle]{lstm_update} which exist only in framework-agnostic form remain unchanged.

Before overwriting, the original framework-agnostic \lstinline[style=mystyle]{__dict__} is saved internally, ready for whenever the specific framework is unset again. So long as a framework is set globally, the local framework specification mechanisms are unavailable. This is because the framework-specific functions do not accept an \lstinline[style=mystyle]{f} argument or call \lstinline[style=mystyle]{get_framework(*args, f=f)}, as the framework agnostic functions do.

\subsection{Ivy Classes}

While Ivy is a fully functional framework at it's core, Ivy does also provide some classes which build directly on these functional primitives. Common trainable neural network layers and stateful optimizers are provided, which derive from the base classes \lstinline[style=mystyle]{ivy.Module} and \lstinline[style=mystyle]{ivy.Optimizer} respectively. Importantly, these high level classes do not abstract high-level classes from the backend frameworks. New classes are built directly on top of Ivy's framework-agnostic functional API, making the new classes stable, maintainable and easily extensible. The details of these layers and optimizers are beyond the scope of this paper, which instead focuses on the framework's functional foundation and it's libraries.
\section{Ivy Libraries}
\label{sec:ivy_libraries}

% \begin{figure*}[h!]
% \centering
% \includegraphics[width=\textwidth]{figures/ivy_usage_across_libs.png}
%   \caption{Usages of core Ivy functions across the four Ivy libraries.}
%   \label{fig:ivy_usage_across_libs}
% \end{figure*}

By building on Ivy's core framework-agnostic API, many high-level framework-agnostic Ivy libraries are possible. We provide an initial set of libraries for mechanics, 3D vision, robotics, and differentiable RL environments. Like the core API, every function in these libraries are unit tested, and all support arbitrary batch dimensions of the inputs. We provide brief overviews of these four libraries below. To offer an insight into which Ivy functions are useful for creating which libraries, the frequencies of Ivy core functions used for each library are presented in Appendix \ref{app:ivy_usage_in_libs}.

\paragraph{Ivy Mech} provides functions for conversions of orientation, pose, and positional representations, as well as frame-of-reference transformations, and other more applied functions.

\paragraph{Ivy Vision} focuses predominantly on 3D vision, with functions for camera geometry, image projections, co-ordinate frame transformations, forward warping, inverse warping, optical flow, depth triangulation, voxel grids, point clouds, implicit rendering and signed distance functions.

\paragraph{Ivy Robot} provides functions and classes for gradient-based motion planning and trajectory optimization. Classes are provided both for mobile robots and robot manipulators.

\paragraph{Ivy Gym} provides differentiable implementations of the classic control tasks from OpenAI Gym. The differentiable nature of the environments means that the cumulative reward can be directly optimized for in a supervised manner, without need for reinforcement learning.

The functions in these libraries can all be integrated directly into arbitrary computation graphs for end-to-end gradient-based learning. We consider an end-to-end example using these libraries in Section \ref{sec:end-to-end}.
\section{A Spectrum of Users}
\label{sec:spectrum_of_users}

\begin{figure}[h!]
\centering
\includegraphics[width=0.45\textwidth]{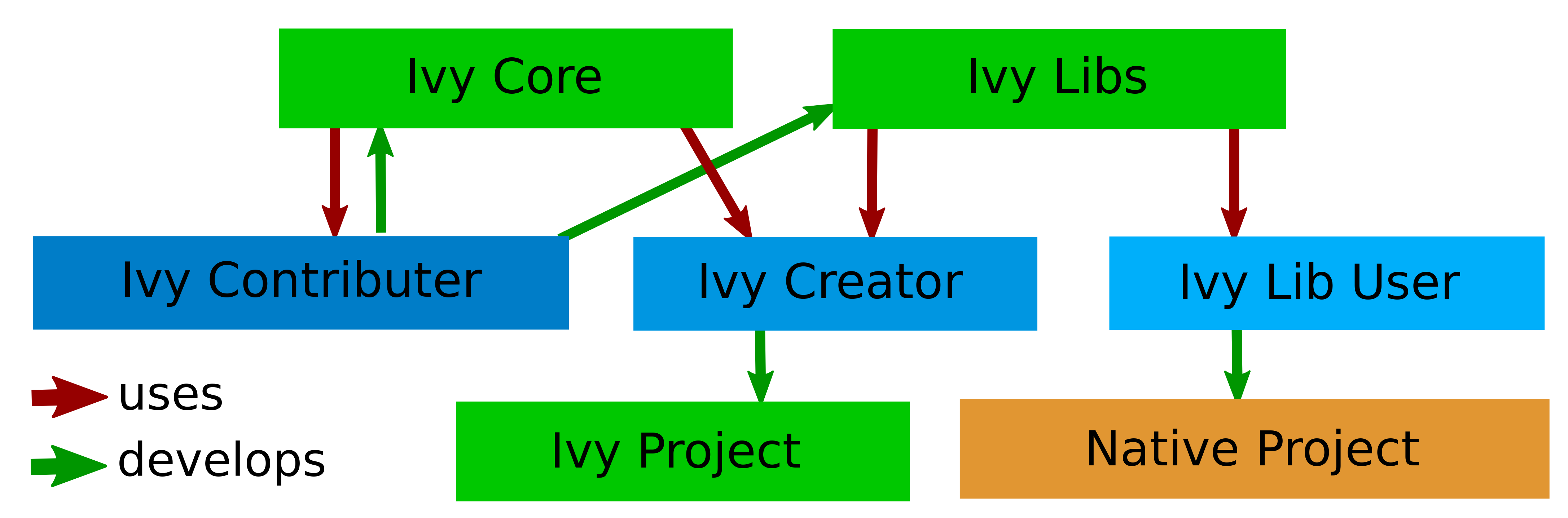}
  \caption{A spectrum of Ivy users.}
  \label{fig:spectrum_of_users}
\end{figure}

Ivy can be used in a variety of ways, depending on the needs and goals of the user. We consider three different hypothetical groups of Ivy users: Ivy contributors, Ivy creators and Ivy library users. We also show how these groups fall onto a broader spectrum of potential users, see Fig \ref{fig:spectrum_of_users}.

\paragraph{Ivy Contributors} exist on one end of the spectrum. If a developer would like to release their own applied DL library, and do so in a manner that maximizes the number of potential users across different frameworks, then writing their library in Ivy provides the solution. An Ivy contributor \textit{uses} Ivy Core to \textit{develop} an Ivy library, potentially helping further \textit{develop} Ivy Core in the process.

\paragraph{Ivy Library Users} exist on the other end of the spectrum. This is likely the most common Ivy user, who simply uses the existing Ivy libraries to supplement their own projects in their own preferred native framework. For example, a TensorFlow user working on DL for computer vision might just want to use some of the Ivy vision functions in their own project. An Ivy library user therefore \textit{uses} Ivy libraries to \textit{develop} their own native project. A code example for this type of user is provided in Section \ref{sec:gdp}.

\paragraph{Ivy Creators} exist somewhat in the middle of the spectrum. They do not explicitly contribute to Ivy with the creation of new Ivy libraries, but they also do more than just copy existing functions into their native project. An Ivy creator uses both Ivy core and the Ivy libraries to implement substantial parts of their own personal project in Ivy. Once this project is released online, the project can be forked by others and integrated directly into new projects, regardless of their frameworks. This then maximizes the direct audience of the code. An example of an Ivy creator's pure-Ivy trainable fully connected network is shown below.

% \begin{minipage}{\linewidth}
\begin{lstlisting}[
style=mystyle,
label=ivy_creator_code,
breakindent=40pt]
class IvyFcModel(ivy.Module):

  def __init__(self):
    self.linear0 = ivy.Linear(3, 64)
    self.linear2 = ivy.Linear(64, 1)
    ivy.Module.__init__(self)

  def _forward(self, x):
    x = ivy.relu(self.linear0(x))
    return ivy.sigmoid(self.linear2(x))
\end{lstlisting}
% \end{minipage}

The network can then be trained using Ivy as shown below.

% \begin{minipage}{\linewidth}
\begin{lstlisting}[
style=mystyle,
label=ivy_creator_code,
breakindent=40pt]
ivy.set_framework('torch')
model = IvyFcModel()
optimizer = ivy.Adam(1e-4)
x_in = ivy.array([1., 2., 3.])

def loss_fn(v):
  return ivy.reduce_mean(model(x_in, v=v))

for step in range(100):
  loss, grads = ivy.execute_with_gradients(loss_fn, model.v)
  model.v = optimizer.step(model.v, grads)
\end{lstlisting}
% \end{minipage}

Alternatively, the network can be used as a parent class alongside a framework-specific model parent class to create a framework-specific trainable child class. The network can then be trained using the native framework's optimizers and trainers. A code examples is presented in Appendix \ref{app:creator_training_options}.

Combined, these hypothetical user groups form a spectrum of potential Ivy users. Ivy's focus on abstracting only low-level functions, and the ability to build new classes on top of these primitives, make it easy to write Ivy code directly alongside native code. This means the developer stays in complete control regarding the depth of the Ivy abstraction in their own projects, as previously outlined in Fig \ref{fig:abstraction_hierarchy}. This flexibility in Ivy's usage underpins the wide variety in potential Ivy users.
\begin{figure*}[h!]
\centering
\includegraphics[width=\textwidth]{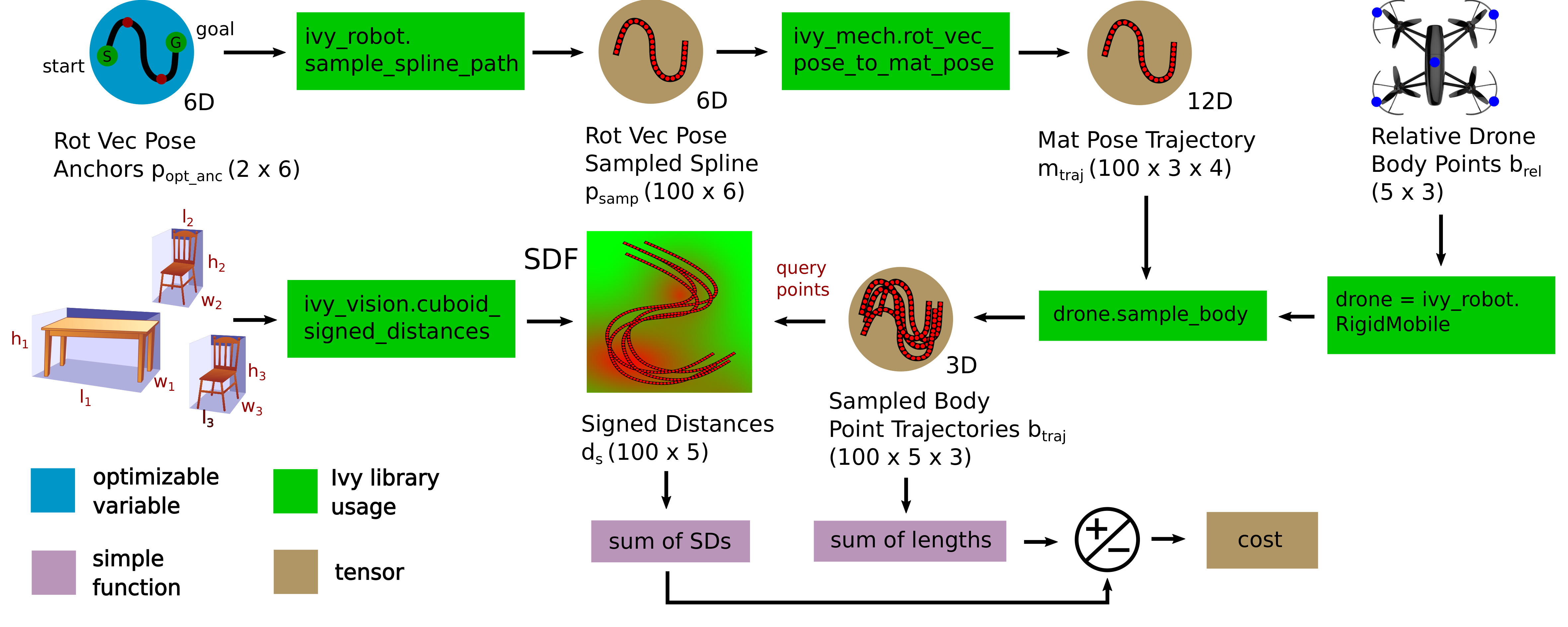}
  \caption{An example application using Ivy. The schema shows how functions from the mechanics, vision and robotics libraries are combined to create a gradient-based motion planning pipeline for a drone, in a cluttered indoor scene.}
  \label{fig:motion_planning_pipeline}
\end{figure*}

\section{End-to-End Integration}\label{sec:end-to-end}

The functions from all Ivy libraries can be integrated into arbitrary computation graphs, such as neural networks, for gradient-based end-to-end training. This is useful for many areas of intersectional research, which explore the integration of conventional parameter-free computation within neural-network based deep learning. The libraries are also applicable to gradient-based methods outside of deep learning. We explore once such example in this section, which combines the different Ivy libraries in an intersectional application.

Specifically, we explore the combined application of the mechanics, vision and robotics libraries to gradient-based motion planning of a drone in a scene with obstacles, see Fig \ref{fig:motion_planning_pipeline}. This takes on a similar formulation to a variety of existing works \cite{ratliff2009chomp, schulman2014motion}. The full code for this example is given in Appendix \ref{app:motion_planning_code}.

First, we define a start pose $p_s\in\mathbb{R}^6$ and target pose $p_t\in\mathbb{R}^6$ for the drone in the scene, both represented as a cartesian position and rotation vector. We then define two intermediate optimizable pose anchor points $p_{opt~anc}\in\mathbb{R}^{2\times6}$. Combined, these represent the four anchor points of a spline $p_{anc}\in\mathbb{R}^{4\times6}$.

The spline is then interpolated and sampled using method \lstinline[style=mystyle]{ivy_robot.sample_spline_path}, returning a more dense trajectory of poses from start to goal, $p_{samp}\in\mathbb{R}^{100\times6}$. The method \lstinline[style=mystyle]{ivy_mech.rot_vec_pose_to_mat_pose} is then used to convert this into a trajectory of pose matrices $m_{traj}\in\mathbb{R}^{100\times3\times4}$.

An \lstinline[style=mystyle]{ivy_robot.RigidMobile} class is also instantiated as a \lstinline[style=mystyle]{drone} object, receiving a collection of 5 relative body points $b_{rel}\in\mathbb{R}^{5\times3}$ in the constructor. In this example, the points represent the centroid and the four outer corners of the drone, but the class enables arbitrary rigid mobile robots. The public method \lstinline[style=mystyle]{drone.sample_body} is then called, receiving the trajectory of matrix poses $m_{traj}$, to produce body point trajectories $b_{traj}\in\mathbb{R}^{100\times5\times3}$ in world space.

The scene is represented as a collection of bounding boxes, one for each object, and the method \lstinline[style=mystyle]{ivy_vision.cuboid_signed_distances} is used to convert this scene description into a single scene-wide signed distance function (SDF).
This SDF is then queried using the body point trajectories $b_{traj}$ and summed, the lengths of each trajectory in $b_{traj}$ are also summed, and the sum of lengths and negative sum of signed-distances are combined to create the motion planning cost function.

\begin{figure}[h!]
\centering
\includegraphics[width=0.48\textwidth]{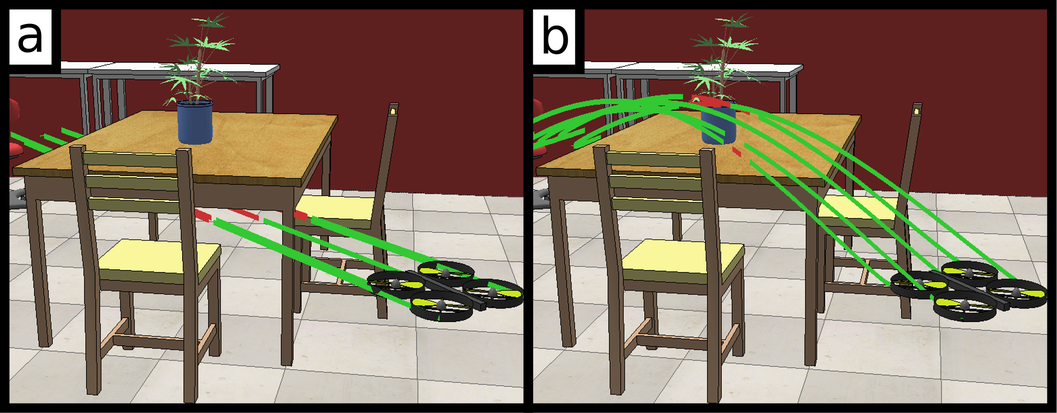}
  \caption{Application of gradient-based motion planning for a drone in a scene with obstacles. (a) path from start to goal at initialization, green shows regions of positive SDF, and red shows negative, which correspond to colliding points. (b) the same path after a few iterations of gradient descent, the path is still not yet collision free with respect to the object bounding boxes, as seen by some small segments of the path which remain red.}
  \label{fig:drone_path}
\end{figure}

The code provided in Appendix \ref{app:motion_planning_code} is a simplified version of an interactive demo provided in the robotics library. Scene renderings at various stages of this interactive demo are provided in Fig \ref{fig:drone_path}. For visualization and simulation we use PyRep \cite{james2019pyrep} and CoppeliaSim \cite{rohmer2013v}.

While the Ivy libraries are predominantly targeted at neural-network integration, this demo highlights how the different Ivy libraries can be combined to also enable gradient-based solutions without neural networks.
\section{Framework Evaluations}

As is the case for most software abstractions, the Ivy abstraction brings improvements for development time, at a small expense of runtime. In this section, we first perform a simple line-of-code (LoC) analysis, to assess how Ivy and it's libraries can accelerate rapid prototyping by reducing lines of code. We then perform a runtime analysis of all the functions in Ivy core, to assess the overhead introduced by the wrapping of backend functions, which brings all backend frameworks into syntactic and behavioural alignment.

\subsection{Line of Code Analysis}

There are two mechanisms by which Ivy reduces the lines of code required for developers. Firstly, Ivy makes it possible to write a library once, with joint support of all DL frameworks. Ivy currently supports 5 backend frameworks, which means all Ivy libraries use only $20\%$ of the code that would be required compared to the alternative of creating framework-specific libraries. Secondly, the Ivy libraries offer a variety of commonly used functions in different areas of applied DL. This avoids the need for Ivy users to implement these functions themselves, reducing lines of code in their own projects.

To quantify these points with a concrete example, we analyse the lines of code required to implement the motion planning pipeline from Sec \ref{sec:end-to-end}, both with and without Ivy and it's libraries. We consider the lines of code required from the perspective of the Ivy user, wishing to implement this demo in a manner that supports all frameworks.

We first assume access to both Ivy and it's libraries, which results in 100 LoC. These are provided in Appendix \ref{app:motion_planning_code}.

We next assume that the libraries do still exist, but Ivy does not exist, and so we assume the libraries are implemented in each of the native frameworks PyTorch, TensorFlow, JAX, and MXNet. This would mean four separate motion planning demo scripts would be required in order to support all frameworks, bringing the total LoC to $100 \times 4 = 400$. Numpy is not included in this case, as it does not support automatic gradients, which are required for this demo.

We next consider the LoC assuming that Ivy does exist, but the Ivy libraries do not exist. Table \ref{table:func_LoC} quantifies the LoC for each of the functions used in the example from Section \ref{sec:end-to-end}, outlined in Figure \ref{fig:motion_planning_pipeline}.

\begin{table}[h!]
\begin{center}
 \begin{tabular}{|| l || c ||} 
 \hline
 \lstinline[style=mystyle]|ivy_robot.RigidMobile| & 53 \\
 \hline
 \lstinline[style=mystyle]|ivy_robot.sample_spline_path| & 133 \\
 \hline
 \lstinline[style=mystyle]|ivy_mech.rot_vec_pose_to_mat_pose| & 108 \\
 \hline
 \lstinline[style=mystyle]|ivy_vision.cuboid_signed_distances| & 61 \\
 \hline
\end{tabular}
\caption{Lines of code for the different Ivy library functions used in the motion planning example from Section \ref{sec:end-to-end}.}
\label{table:func_LoC}
\end{center}
\end{table}

Therefore, without the existence of the Ivy libraries, each function would need to be implemented as part of the demo, and the total demo LoC increases to $100 + 53 + 133 + 108 + 61 = 455$.

Finally, we consider the case where neither Ivy nor the Ivy libraries exist. Taking the previous result for no Ivy libraries $455$ as a starting point, the demo would now also need to be repeated for each specific framework, bringing the total LoC to $455 \times 4 = 1820$. All of these results are summarized in Table \ref{table:LoC_results}.

\begin{table}[h!]
\begin{center}
\resizebox{\columnwidth}{!}{%
 \begin{tabular}{|| c | c || c | c || c | c || c | c ||} 
 \hline
 \multicolumn{2}{||c||}{Naive} & \multicolumn{2}{c||}{Ivy Only} & \multicolumn{2}{c||}{Ivy Libs Only} & \multicolumn{2}{c||}{Ivy and Libs} \\ 
 \hline
 LoC & \% & LoC & \% & LoC & \% & LoC & \% \\
 \hline
 \hline
  1820 & 100 & 455 & 25 & 400 & 22 & 100 & 5 \\
 \hline
\end{tabular}%
}
\caption{Lines of code to implement the demo in Section \ref{sec:end-to-end}, for varying availability of both Ivy and the Ivy libraries.}
\label{table:LoC_results}
\end{center}
\end{table}

As can be seen in Table \ref{table:LoC_results}, the demo only requires $\sim5\%$ of the LoC compared to implementing the same demo without using Ivy or it's libraries, in a manner that supports all frameworks. Of course, one could argue that this example is somewhat contrived, with the example being specifically chosen to maximally utilize the libraries. It is indeed true that many useful functions do not yet exist in the Ivy libraries, and these would then need to be implemented in local project codebases, thus increasing LoC.

However, if many such functions become apparent to developers, then these functions can be added to the Ivy libraries, enabling more LoC reductions for future users of the libraries. Overall, this motion planning demo exemplifies the dramatic LoC reduction which is possible when using Ivy and the Ivy libraries to create framework-agnostic code.

\begin{figure*}[h!]
\centering
\includegraphics[width=\textwidth]{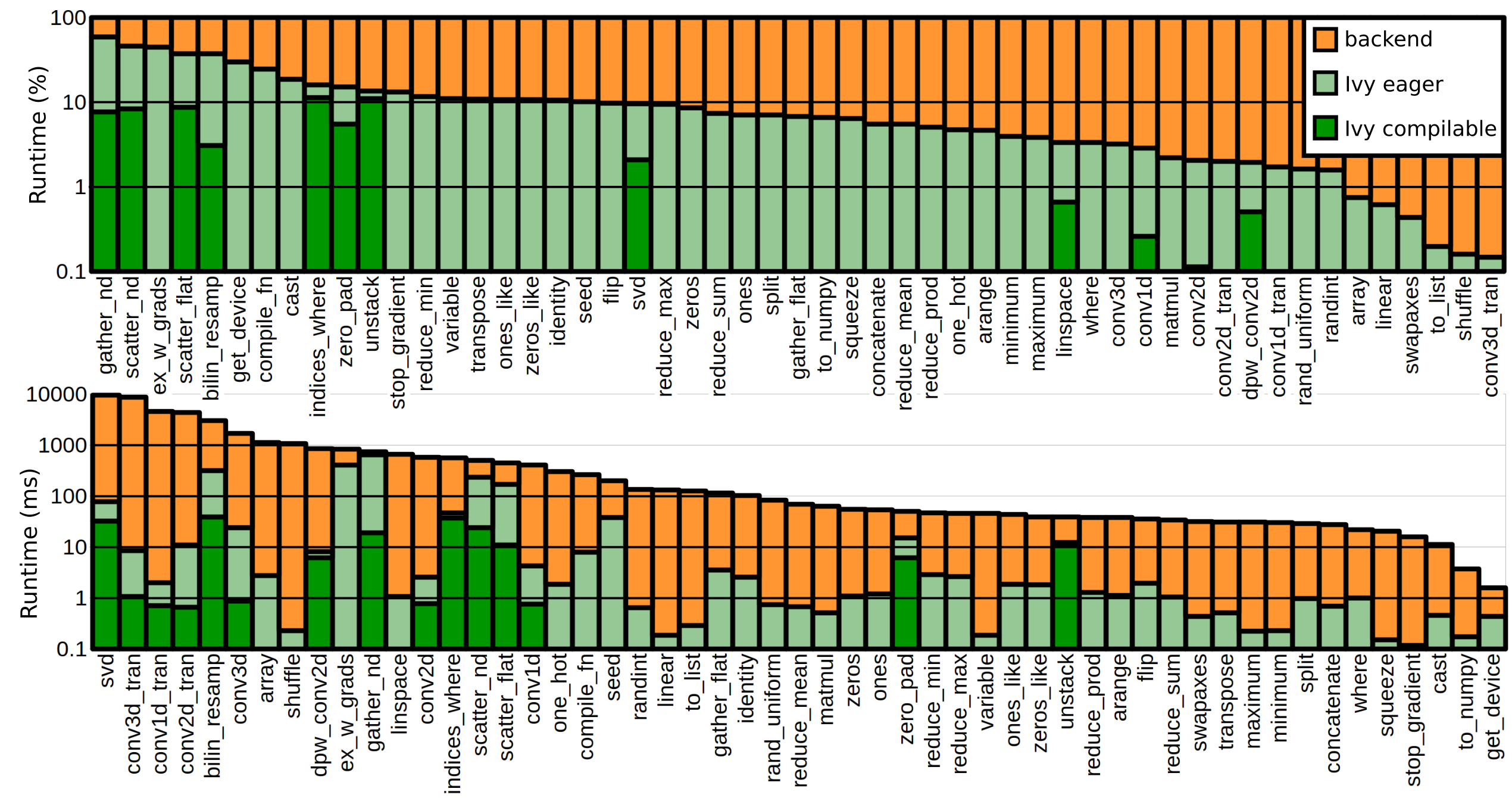}
  \caption{Runtimes for each Ivy core method which exhibits some Ivy overhead. The bars are cumulative, with the colors representing the proportion of the runtime consumed by each of the 3 code groups, explained in Section \ref{sec:RuntimeAnalysis}. Note the log scale in both plots.}
  \label{fig:runtime_analysis}
\end{figure*}

\subsection{Ivy Core Runtime Analysis}
\label{sec:RuntimeAnalysis}

In order to assess the overhead introduced by the Ivy abstraction, we perform a runtime analysis for each core function using all possible backend frameworks, and assess how much inference time is consumed by the Ivy abstraction in both eager mode and compiled mode. Ivy code can be compiled using \lstinline[style=mystyle]{ivy.compile_fn()}, which wraps the compilation tools from the native framework. Our analysis only considers 53 of the 101 core functions implemented at the time of writing, as the remaining 48 Ivy functions incur no overhead for any of the backend frameworks. Each function is called with all inputs of flattened size $1\times10^4$ where possible, and is executed on the CPU.

To perform this analysis, we separate the lines of code for each Ivy function into 3 code groups: (a) backend, (b) Ivy compilable and (c) Ivy eager. Backend code refers to the native tensor operation or operations being abstracted by Ivy. These operations form part of the compilable computation graph. Ivy compilable refers to overhead tensor operations which also form part of the compilable computation graph. A good example is reshape and transpose operations which are sometimes required to unify input-output tensor shapes between frameworks. Finally, Ivy eager refers to Ivy overhead which is only executed when running the backend framework in eager execution mode. If compiled, this code is not run as part of the graph. Examples include inferring the shapes of input tensors via the \lstinline[style=mystyle]{.shape} attribute, inferring data-types from string input, and constructing new shapes or transpose indices as lists, for defining tensor operations which themselves form part of the compilable computation graph. A function which consists of \textcolor{backendcolor}{backend} and \textcolor{ivycolor}{Ivy compilable} code is presented below. The transpose operation is necessary to return the output in the expected format.

\begin{lstlisting}[
style=mystyle,
label=torch_svd_code_groups,
linebackgroundcolor={
\ifnum\value{lstnumber}=1\color{backcolor}\fi
\ifnum\value{lstnumber}=2\color{backendbackcolor}\fi
\ifnum\value{lstnumber}=3\color{ivycompilablebackcolor}\fi
\ifnum\value{lstnumber}=4\color{backcolor}\fi}
]
def svd(x, batch_shape=None):
  U, D, V = torch.svd(x)
  VT = torch.transpose(V, -2, -1)
  return U, D, VT
\end{lstlisting}

A function which consists of \textcolor{backendcolor}{backend} and \textcolor{ivyeagercolor}{Ivy eager} code is presented below. The dictionary lookup is not compiled into the computation graph, and is only run on the first function call which is responsible for compiling the graph.

\begin{lstlisting}[
style=mystyle,
label=torch_identity_code_groups,
linebackgroundcolor={
\ifnum\value{lstnumber}=1\color{backcolor}\fi
\ifnum\value{lstnumber}=2\color{ivyeagerbackcolor}\fi
\ifnum\value{lstnumber}=3\color{backendbackcolor}\fi}
]
def cast(x, dtype_str):
  dtype_val = torch.__dict__[dtype_str]
  return x.type(dtype_val)
\end{lstlisting}

\begin{table*}[t]
\begin{center}
\resizebox{2\columnwidth}{!}{%
 \begin{tabular}{|| c || c | c || c | c || c | c || c | c || c || c | c ||} 
 \cline{2-12}
 \multicolumn{1}{c||}{} & \multicolumn{2}{c||}{JAX} & \multicolumn{2}{c||}{TensorFlow} & \multicolumn{2}{c||}{PyTorch} & \multicolumn{2}{c||}{MXNet} & \multicolumn{1}{c||}{NumPy} & \multicolumn{2}{c||}{Mean} \\
 \cline{2-12}
  \multicolumn{1}{c||}{} & eager & compiled & eager & compiled & eager & compiled & eager & compiled & eager & eager & compiled \\
 \hline
 \hline
 Ivy Mech & 0.1 & 0.0 & 0.3 & 0.0 & 0.4 & 0.0 & 1.5 & 0.0 & 0.2 & 0.4 & 0.0 \\
 \hline
 Ivy Vision & 12.1 & 0.5 & 2.4 & 0.4 & 5.6 & 0.3 & 10.8 & 2.0 & 1.2 & 6.3 & 0.4 \\
 \hline
 Ivy Robot & 0.4 & 0.0 & 0.7 & 0.0 & 0.3 & 0.0 & 2.5 & 0.0 & 1.0 & 0.7 & 0.0 \\
 \hline
 Ivy Gym & 0.4 & 0.0 & 0.6 & 0.0 & 0.9 & 0.0 & 3.1 & 0.0 & 0.6 & 0.8 & 0.0 \\
 \hline
 \hline 
 Mean & 3.25 & 0.1 & 1.0 & 0.1 & 1.8 & 0.1 & 4.5 & 0.5 & 1.1 & 2.0 & 0.1 \\
 \hline
\end{tabular}%
}
\caption{Percentage slowdown when using Ivy in either eager or compiled mode with each of the Ivy libraries, using each of the possible backend frameworks.}
\label{table:library_runtime_analysis}
\end{center}
\end{table*}

In order to simplify the runtime analysis, we time all Ivy functions only in eager mode, by using the method \lstinline[style=mystyle]{time.perf_counter()} from the \lstinline[style=mystyle]{time} module between adjacent code groups. While the absolute runtimes of eager functions will be slower than compiled functions, we find that the \textit{relative} runtimes between different tensor operations in eager mode is a good approximation for their \textit{relative} runtimes in compiled mode. Our analysis focuses on the proportionate overhead of Ivy, and not the absolute compiled runtimes, and so this approximation is still informative for our analysis. The runtime analysis for each core function averaged across the backend frameworks is presented in Figure \ref{fig:runtime_analysis}, and framework-specific runtimes are presented in Appendix \ref{app:runtime_analysis}.

Finally, by combining the method usage frequencies for each library (see Appendix \ref{app:ivy_usage_in_libs}) with the Ivy overhead runtimes, we assess the Ivy overhead when using each of the four Ivy libraries in both eager mode and compiled mode. We compute these values separately for each backend framework. The results are presented in Table \ref{table:library_runtime_analysis}.

Overall, we can see that the overhead is very minimal both when compiling Ivy code and when running in eager execution mode. We can also see that the vision library incurs the largest Ivy overhead. This is due to the frequent usage of gather and scatter functions for rendering. The ``overhead'' in the graph for these functions are related to extensions over the simpler backend methods, with added support for handling multiple dimensions. However, we do not formally distinguish between ``overhead'' and ``extensions'' in our analysis, as the boundary between these is difficult to determine objectively. Even without this distinction, the measured Ivy overhead is very minimal in most cases.
\section{Conclusion and Future Work}

In this paper we present Ivy, a templated deep learning framework supporting TensorFlow, PyTorch, MXNet, Jax, and Numpy. Ivy offers the potential for creating framework-agnostic DL libraries, which are usable in both present and hypothetical future frameworks. We provide four initial Ivy libraries for mechanics, 3D vision, robotics, and differentiable environments. We welcome developers to join the Ivy community by writing their own functions, layers and libraries in Ivy, maximizing their direct audience and helping to accelerate DL research through the creation of ``lifelong'' inter-framework codebases.

Regarding the future vision for Ivy, we will continue extending the derived libraries, as well as adding new libraries for additional research fields. We also will continue developing Ivy Core, to remain compatible with all the latest DL framework developments, and add support for new Python frameworks as and when they arrive. We will strive to support the community of open DL research through our framework, and continue to encourage collaboration and contributions from the community.

\section*{Acknowledgements}

We are grateful to many individuals for providing helpful feedback on the Ivy paper, code-base and the broader Ivy project. Specifically, we would like to thank Martin Asenov, Patrick Bardow, Michael Bloesch, Chris Choi, Jan Czarnowski, Andrew Davison, Ankur Handa, Dorian Hennings, Edward Johns, Tristan Laidlow, Zoe Landgraf, Stefan Leutenegger, Wenbin Li, Shikun Liu, Robert Lukierski, Hide Matsuki, Andrea Nicastro, Joe Ortiz, Sajad Saeedi, Edgar Sucar, Dimos Tzoumanikas, Kentaro Wada, Binbin Xu, and Shuaifeng Zhi for helpful comments and feedback.

% \textbf{Do not} include acknowledgements in the initial version of
% the paper submitted for blind review.

% If a paper is accepted, the final camera-ready version can (and
% probably should) include acknowledgements. In this case, please
% place such acknowledgements in an unnumbered section at the
% end of the paper. Typically, this will include thanks to reviewers
% who gave useful comments, to colleagues who contributed to the ideas,
% and to funding agencies and corporate sponsors that provided financial


%\bibliography{ref}

\bibliographystyle{mlsys2020}

\onecolumn
\appendix
\section{Appendices}

\subsection{Ivy Usage in Libraries}
\label{app:ivy_usage_in_libs}

\begin{figure}[H]
\centering
\includegraphics[width=.9\textwidth]{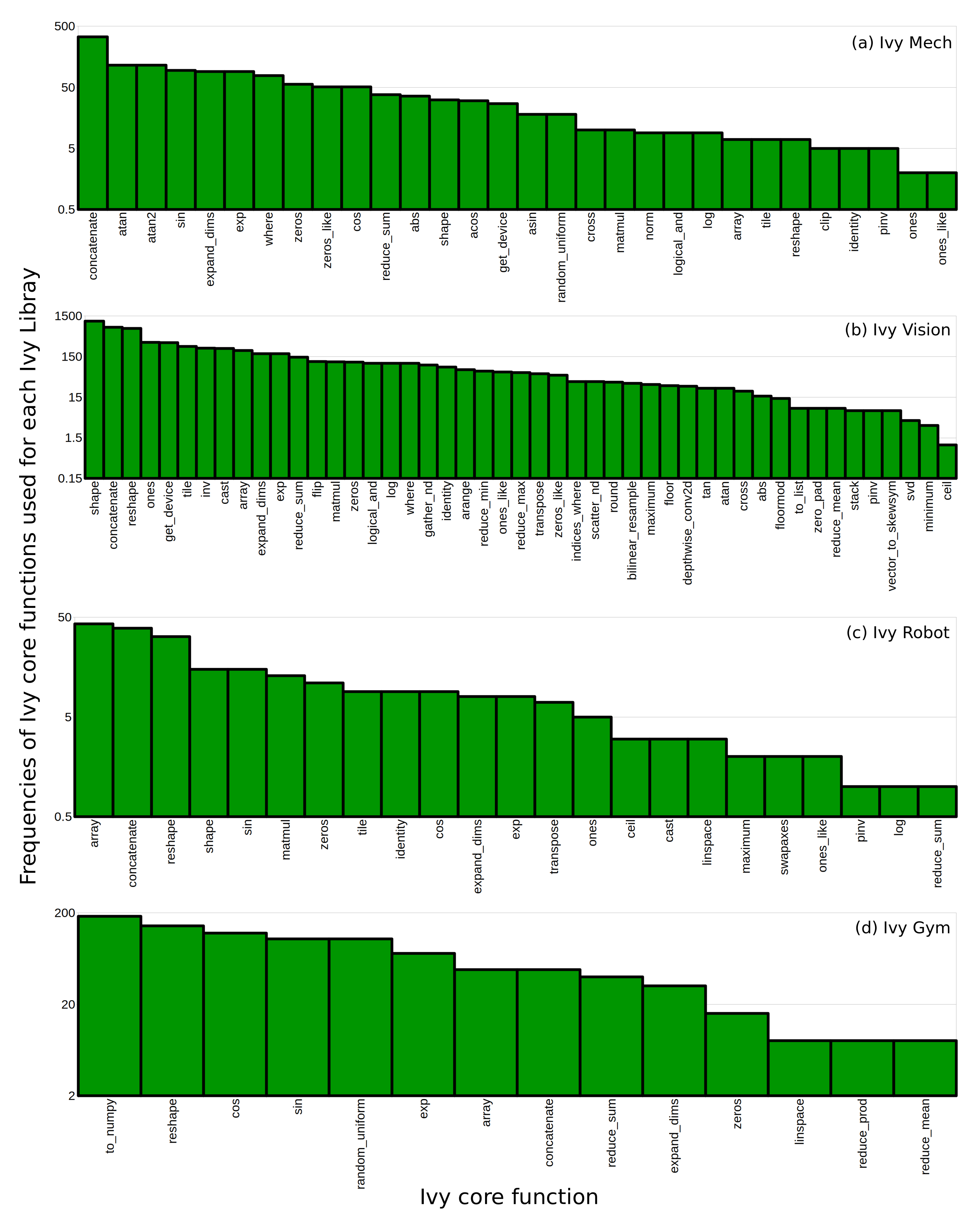}
  \caption{Usages of core Ivy functions in each of the four Ivy libraries.}
  \label{fig:ivy_usage_in_libs}
\end{figure}

The frequency of Ivy core functions appearing in each of the four Ivy libraries is presented in Figure \ref{fig:ivy_usage_in_libs}.

\subsection{Ivy Training Options}
\label{app:creator_training_options}

If an Ivy user intends to create a trainable model, then that model can either be trained using a pure Ivy pipeline, or trained directly in one of the native frameworks, using native trainer and optimizer classes. First, we recap the simple fully connected model outlined in Section \ref{sec:spectrum_of_users}.

% \begin{minipage}{\linewidth}
\begin{lstlisting}[
style=mystyle,
label=ivy_creator_code,
breakindent=40pt]
class IvyFcModel(ivy.Module):

  def __init__(self):
    self.linear0 = ivy.Linear(3, 64)
    self.linear2 = ivy.Linear(64, 1)
    ivy.Module.__init__(self)

  def _forward(self, x):
    x = ivy.relu(self.linear0(x))
    return ivy.sigmoid(self.linear2(x))
\end{lstlisting}
% \end{minipage}

We next recap the pure-Ivy training pipeline, outlined in section \ref{sec:spectrum_of_users}.

% \begin{minipage}{\linewidth}
\begin{lstlisting}[
style=mystyle,
label=ivy_creator_code,
breakindent=40pt]
ivy.set_framework('torch')
model = IvyFcModel()
optimizer = ivy.Adam(1e-4)
x_in = ivy.array([1., 2., 3.])

def loss_fn(v):
  return ivy.reduce_mean(model(x_in, v=v))

for step in range(100):
  loss, grads = ivy.execute_with_gradients(loss_fn, model.v)
  model.v = optimizer.step(model.v, grads)
\end{lstlisting}
% \end{minipage}

Alternatively, the network can be used as a parent class alongside a framework-specific parent class to create a framework-specific trainable child class. This enables the network to be trained using the native framework's own optimizers and trainers, like so:

% \begin{minipage}{\linewidth}
\begin{lstlisting}[
style=mystyle,
label=ivy_creator_train_native_code,
breakindent=40pt]
class TorchFcModel(torch.nn.Module, IvyFcModel):

  def __init__(self):
    torch.nn.Module.__init__(self)
    IvyFcModel.__init__(self)
    self._assign_variables()

  def _assign_variables(self):
    self.v.map(
      lambda x, kc: self.register_parameter(name=kc, param=torch.nn.Parameter(x)))
      self.v = self.v.map(lambda x, kc: self._parameters[kc])

  def forward(self, x):
    return self._forward(x)
    
ivy.set_framework('torch')
model = TorchFcModel()
optimizer = torch.optim.Adam(model.parameters(), lr=1e-4)
x_in = torch.tensor([1., 2., 3.])

def loss_fn():
  return torch.mean(model(x_in))

for step in range(100):
  loss = loss_fn()
  loss.backward()
  optimizer.step()
\end{lstlisting}
% \end{minipage}

\subsection{Motion Planning Code}
\label{app:motion_planning_code}

The full 100 lines of code for the motion planning demo are provided below. This is a simplified variant of the drone motion planning demo available in the Ivy Robot open source repository. The only difference between the 100 lines of code below and the interactive demo is the lack of integration with a real running simulator, and lack of visualization.

\begin{lstlisting}[
style=mystyle,
label=motion_planning_code]
import ivy
import ivy_mech
import ivy_robot
import ivy_vision


def compute_length(query_vals):
  start_vals = query_vals[0:-1]
  end_vals = query_vals[1:]
  dists_sqrd = ivy.maximum((end_vals - start_vals)**2, 1e-12)
  distances = ivy.reduce_sum(dists_sqrd, -1)**0.5
  return ivy.reduce_sum(distances)


def compute_cost_and_sdfs(learnable_anchor_vals, anchor_points, start_anchor_val,
                          end_anchor_val, query_points, ivy_drone, sdf):
  anchor_vals = ivy.concatenate((ivy.expand_dims(start_anchor_val, 0),
                                 learnable_anchor_vals,
                                 ivy.expand_dims(end_anchor_val, 0)), 0)
  poses = ivy_robot.sample_spline_path(anchor_points, anchor_vals, query_points)
  inv_ext_mat_query_vals = ivy_mech.rot_vec_pose_to_mat_pose(poses)
  body_positions = ivy.transpose(ivy_drone.sample_body(inv_ext_mat_query_vals), (1, 0, 2))
  length_cost = compute_length(body_positions)
  sdf_vals = sdf(ivy.reshape(body_positions, (-1, 3)))
  coll_cost = -ivy.reduce_mean(sdf_vals)
  total_cost = length_cost + coll_cost * 10
  return total_cost, poses, body_positions, ivy.reshape(sdf_vals, (-1, 100, 1))


if __name__ == '__main__':
  # config
  ivy.set_framework('torch')  # change to your backend
  lr = 0.01
  num_anchors = 2
  num_sample_points = 100
  drone_start_pose = ivy.array([-1.1500, -1.0280, 0.6000, 0.0000, 0.0000, 0.6981])
  drone_goal_pose = ivy.array([1.0250, 1.1250, 0.6000, 0.0000, 0.0000, 0.6981])

  # ivy drone
  rel_body_points = ivy.array([[0., 0., 0.],
                               [-0.15, -0.15, 0.],
                               [-0.15, 0.15, 0.],
                               [0.15, -0.15, 0.],
                               [0.15, 0.15, 0.]])
  ivy_drone = ivy_robot.RigidMobile(rel_body_points)

  # simplified scene of two chairs, a table and a plant
  cuboid_ext_mats = ivy.array([[[0.00, 1.00, -0.00, 0.03],
                                [-1.00, 0.00, -0.00, -0.60],
                                [-0.00, 0.00, 1.00, -0.45]],
                               [[-1.00, 0.00, -0.00, 0.28],
                                [-0.00, -1.00, 0.00, -0.65],
                                [-0.00, 0.00, 1.00, -0.45]],
                               [[1.00, -0.00, 0.00, -0.30],
                                [0.00, 1.00, -0.00, 0.00],
                                [-0.00, 0.00, 1.00, -0.37]],
                               [[1.00, -0.00, 0.00, -0.17],
                                [0.00, 1.00, 0.00, 0.02],
                                [-0.00, 0.00, 1.00, -1.03]]])
  cuboid_dims = ivy.array([[0.40, 0.45, 0.91],
                           [0.40, 0.45, 0.91],
                           [1.60, 1.10, 0.75],
                           [0.40, 0.40, 0.56]])


  # sdf
  def sdf(query_positions):
    cuboid_sdfs = ivy_vision.cuboid_signed_distances(cuboid_ext_mats, cuboid_dims,
                                                     query_positions)
    return ivy.reduce_min(cuboid_sdfs, -1, keepdims=True)


  # 1D spline points
  anchor_points = ivy.cast(ivy.expand_dims(ivy.linspace(0, 1, 2 + num_anchors), -1),
                           'float32')
  query_points = ivy.cast(ivy.expand_dims(ivy.linspace(0, 1, num_sample_points), -1),
                          'float32')

  # learnable parameters
  learnable_anchor_vals = ivy.variable(ivy.cast(ivy.transpose(ivy.linspace(
    drone_start_pose, drone_goal_pose, 2 + num_anchors)[..., 1:-1], (1, 0)), 'float32'))
  v = ivy.Container({'anchors': learnable_anchor_vals})

  # optimize
  it = 0
  colliding = True
  clearance = 0.1
  
  while colliding:
    total_cost, grads, poses, body_positions, sdf_vals = ivy.execute_with_gradients(
      lambda xs: compute_cost_and_sdfs(xs.anchors, anchor_points, drone_start_pose,
        drone_goal_pose, query_points, ivy_drone, sdf), v)
    min_sdf = ivy.reduce_min(sdf_vals)
    print('iteration {}, cost = {}, min_sdf - clearance = {}'.format(
      it, ivy.to_numpy(total_cost).item(), ivy.to_numpy(min_sdf - clearance).item()))
    colliding = min_sdf < clearance
    v = ivy.gradient_descent_update(v, grads, lr)
    it += 1
    
  print('collision-free path found!')
\end{lstlisting}

\newpage

\subsection{Framework-Specific Runtime Analysis}
\label{app:runtime_analysis}

\begin{figure}[h!]
\centering
\includegraphics[width=\textwidth]{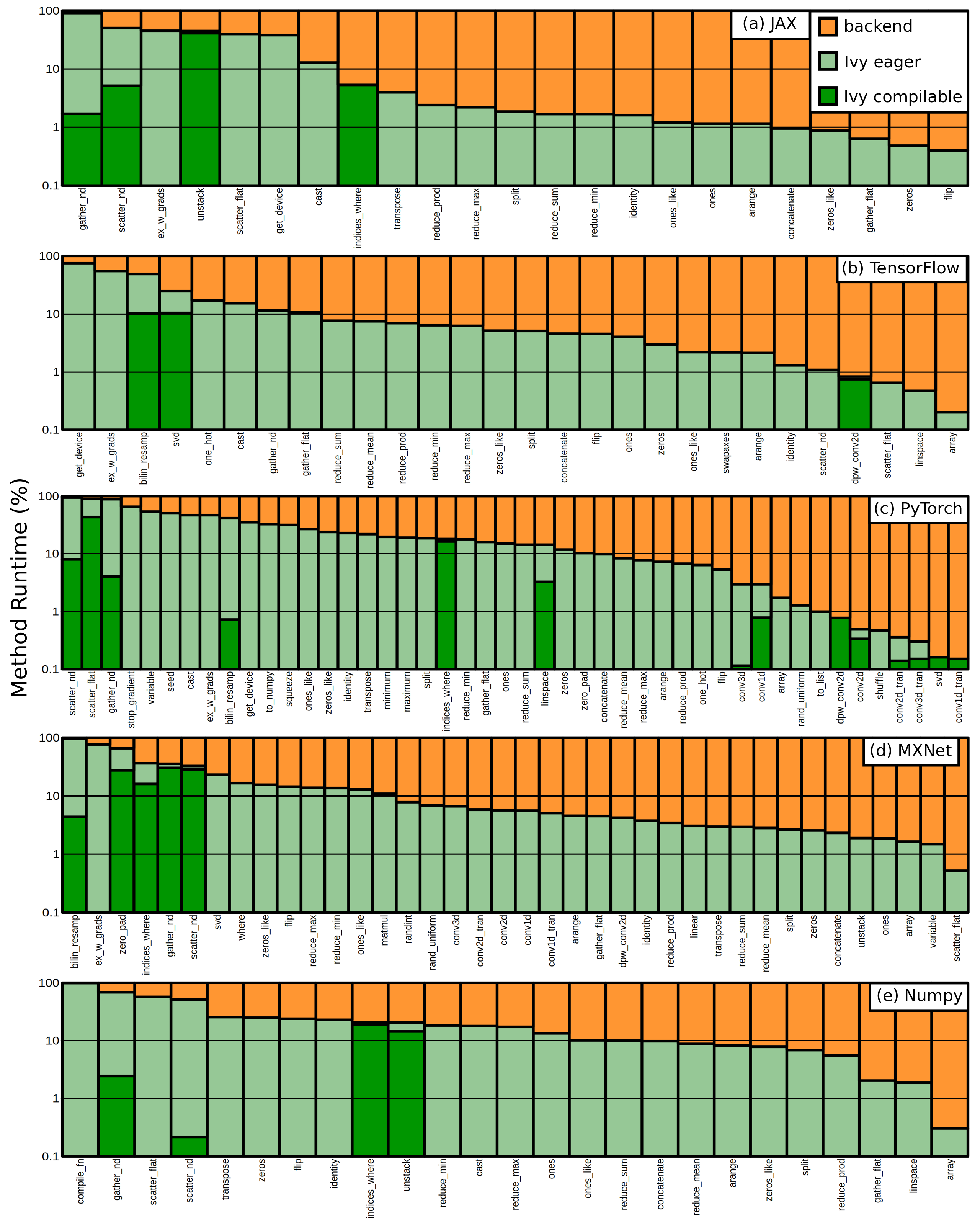}
  \caption{Percentage runtimes for each Ivy core method exhibiting some Ivy overhead, for each specific framework. The bars are cumulative, with colors representing the runtime consumed by each of the 3 code groups outlined in Section \ref{sec:RuntimeAnalysis}. Note the log scale.}
  \label{fig:ratio_runtime_analysis_combined}
\end{figure}

The framework-specific \textit{percentage} runtimes for each Ivy core method which exhibits Ivy overhead, separated into the 3 code groups (a) backend, (b) Ivy compilable and (c) Ivy eager (all explained in Section \ref{sec:RuntimeAnalysis}), are presented in Figure \ref{fig:ratio_runtime_analysis_combined}. The results are presented for each specific backend framework, unlike Figure \ref{fig:runtime_analysis} which provides percentage runtimes averaged across all backend frameworks.

\begin{figure}[h!]
\centering
\includegraphics[width=\textwidth]{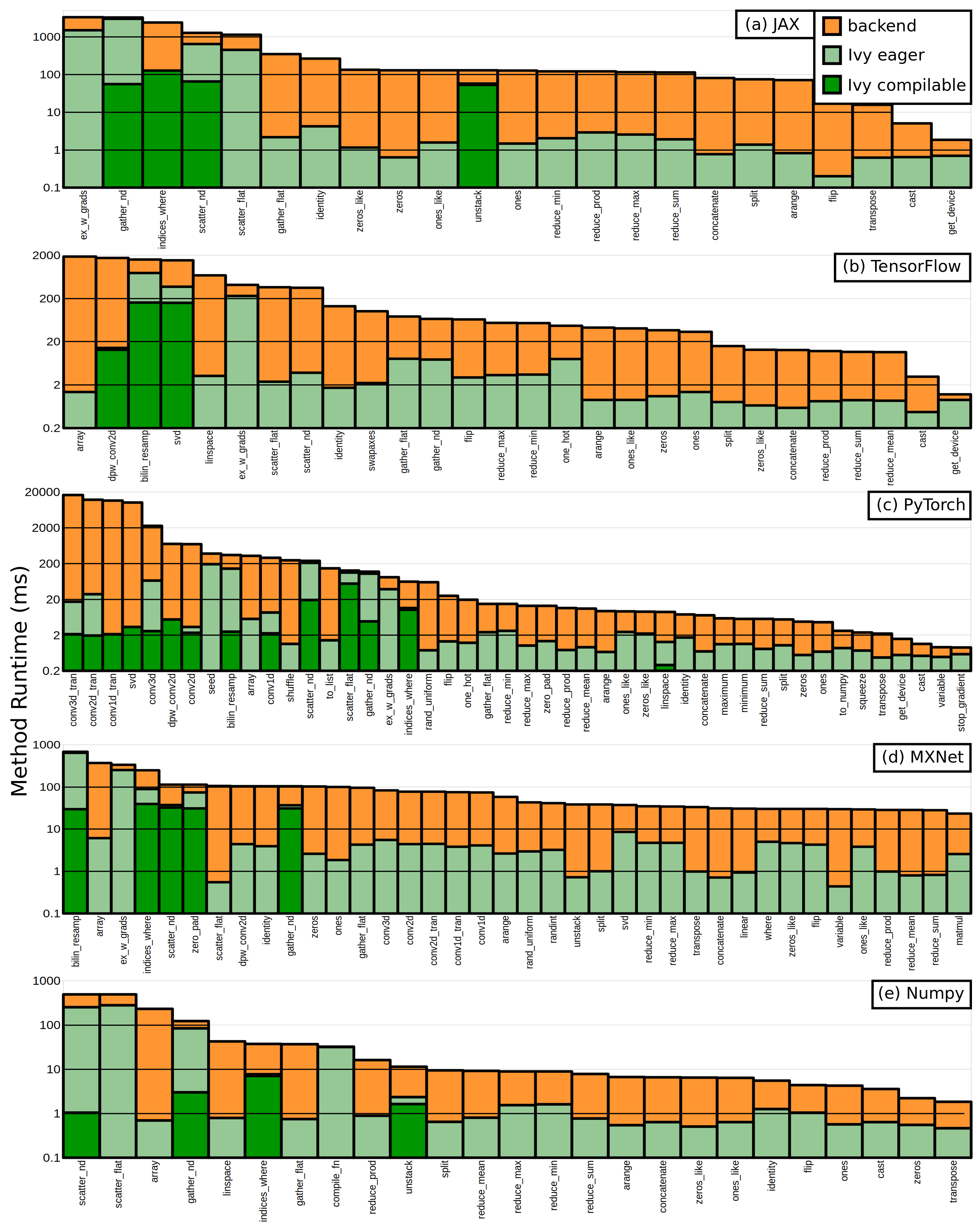}
  \caption{Absolute runtimes for each Ivy core method exhibiting some Ivy overhead, for each specific framework. The bars are cumulative, with colors representing the runtime consumed by each of the 3 code groups outlined in Section \ref{sec:RuntimeAnalysis}. Note the log scale.}
  \label{fig:abs_runtime_analysis_combined}
\end{figure}

The framework-specific \textit{absolute} runtimes for each Ivy core method which exhibits Ivy overhead, separated into the 3 code groups (a) backend, (b) Ivy compilable and (c) Ivy eager (all explained in Section \ref{sec:RuntimeAnalysis}), are presented in Figure \ref{fig:abs_runtime_analysis_combined}. The results are presented for each specific backend framework, unlike Figure \ref{fig:runtime_analysis} which provides absolute runtimes averaged across all backend frameworks.

%%%%%%%%%%%%%%%%%%%%%%%%%%%%%%%%%%%%%%%%%%%%%%%%%%%%%%%%%%%%%%%%%%%%%%%%%%%%%%%
%%%%%%%%%%%%%%%%%%%%%%%%%%%%%%%%%%%%%%%%%%%%%%%%%%%%%%%%%%%%%%%%%%%%%%%%%%%%%%%
% SUPPLEMENTAL CONTENT AS APPENDIX AFTER REFERENCES
%%%%%%%%%%%%%%%%%%%%%%%%%%%%%%%%%%%%%%%%%%%%%%%%%%%%%%%%%%%%%%%%%%%%%%%%%%%%%%%
%%%%%%%%%%%%%%%%%%%%%%%%%%%%%%%%%%%%%%%%%%%%%%%%%%%%%%%%%%%%%%%%%%%%%%%%%%%%%%%
% \appendix
% \section{Please add supplemental material as appendix here}
% %
% Put anything that you might normally include after the references as an appendix here, {\it not in a separate supplementary file}. Upload your final camera-ready as a single pdf, including all appendices.

%%%%%%%%%%%%%%%%%%%%%%%%%%%%%%%%%%%%%%%%%%%%%%%%%%%%%%%%%%%%%%%%%%%%%%%%%%%%%%%
%%%%%%%%%%%%%%%%%%%%%%%%%%%%%%%%%%%%%%%%%%%%%%%%%%%%%%%%%%%%%%%%%%%%%%%%%%%%%%%

\end{document}